\documentclass[runningheads]{llncs}
\newcommand{\prob}{{\cal TextVQG}}
\newcommand{\model}{OLRA}
\usepackage{graphicx}
\usepackage{epigraph}
\usepackage{xcolor}

\begin{document}%

\title{Look, Read and Ask: Learning to Ask \\Questions by Reading Text in Images}
\author{Soumya Jahagirdar\inst{1}\orcidID{0000-0002-3460-9151} \and
Shankar Gangisetty\inst{1}\orcidID{0000-0003-4448-5794} \and
Anand Mishra\inst{2}\orcidID{0000-0002-7806-2557}}

\authorrunning{Jahagirdar et al.}
\institute{KLE Technological University, Hubballi, India\\
\email{\{01fe17bcs212,shankar\}@kletech.ac.in}
\and
Vision, Language, and Learning Group (VL2G)\\
IIT Jodhpur, India\\
\email{mishra@iitj.ac.in}\\
}
\maketitle              
\begin{abstract}
We present a novel problem of text-based visual question generation or \prob{} in short. Given the recent growing interest of the document image analysis community in combining text understanding with conversational artificial intelligence, e.g., text-based visual question answering, \prob{} becomes an important task. \prob{} aims to generate a natural language question for a given input image and an automatically extracted text also known as OCR token from it such that the OCR token is an answer to the generated question. \prob{} is an essential ability for a conversational agent. However, it is challenging as it requires an in-depth understanding of the scene and the ability to semantically bridge the visual content with the text present in the image. To address \prob{}, we present an \underline{O}CR-consistent visual question generation model that \underline{L}ooks into the visual content, \underline{R}eads the scene text, and \underline{A}sks a relevant and meaningful natural language question. We refer to our proposed model as \model{}. We perform an extensive evaluation of \model{} on two public benchmarks and compare them against baselines. Our model -- \model{} automatically generates questions similar to the public text-based visual question answering datasets that were curated manually. Moreover, we significantly outperform baseline approaches on the performance measures popularly used in text generation literature.

\end{abstract}
\section{Introduction}
\epigraph{``To seek truth requires one to ask the right questions."
}{\textit{{Suzy Kassem 
}}}
Developing agents that can communicate with a human has been an active area of research in artificial intelligence (AI). The document image analysis community has also started showing interest in this problem lately as evident from the efforts of the community on text-based visual question answering~\cite{stvqa,docVQA:2021,ocr-vqa,textVQA}, and ICDAR 2019 robust reading challenge on scene text visual question answering~\cite{icdar19comp}. Visual question answering (VQA) is only one desired characteristic of a conversational agent where the agent answers a natural language question about the image. An intelligent agent should also have the ability to ask meaningful and relevant questions with respect to its current visual perception. Given an image, generating meaningful questions, also known as visual question generation (VQG) is an essential component of a conversational agent~\cite{vqg0}. VQG is a precursor of visual dialogue systems and might help in building large-scale VQA datasets automatically. To the best of our knowledge, the document image analysis community has not yet looked into the important problem of VQG leveraging text. In this work, we fill up this gap prevailing in the literature by introducing a novel problem of text-based visual question generation or \prob{} in short.  
\begin{figure}[!t]
\begin{center}
\includegraphics[scale=0.375]{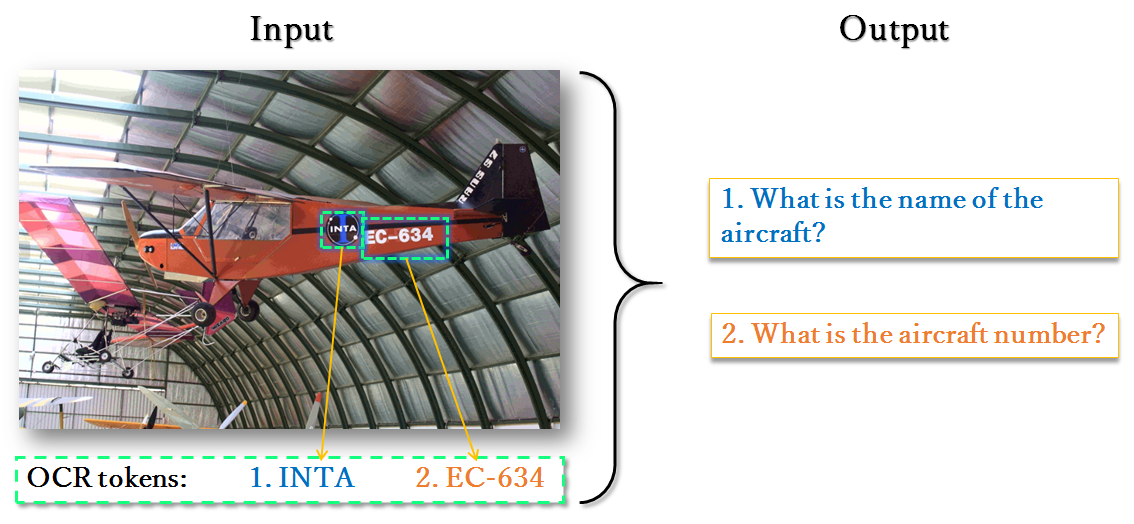}
\caption{\label{fig:goal} \textbf{\prob{}}. We introduce a novel problem of visual question generation by leveraging text in the image. Given an image and an OCR token automatically extracted from it, our goal is to generate a meaningful natural language question whose answer is the OCR token.}
\end{center}
\end{figure}

The \prob{} has the following goal: given a natural image containing text, and an OCR token extracted from it, generate a natural language question whose answer is the OCR token. This problem is challenging because it requires in-depth semantic interpretation of the text as well as the visual content to generate meaningful and relevant questions with respect to the image. VQG has been a well-explored area in vision and language community~\cite{diverse2VQG:2017,vqg-infomax,vqg0,diverse4VQG:2018}. However, these works often ignore the text appearing in the image, and only restrict themselves to the visual content while generating questions. It should be noted that text in the image helps in asking semantically meaningful questions connecting visual and textual content. For example, consider the image shown in Fig.~\ref{fig:goal}. Given an image of an aircraft and two words -- \emph{INTA} and \emph{EC-634} written on it, we aim to automatically generate questions such as ``What is the name of the aircraft?" and ``What is the aircraft number?".  
\newline
\newline
\noindent\textbf{Baselines:} Motivated by the baselines presented in the VQG literature~\cite{vqg0}, a few of the plausible approaches to address \prob{} are as follows: (i) \textbf{maximum-entropy language model (MELM):} which uses the extracted OCR tokens based on their confidence scores along with detected objects to generate question, (ii) \textbf{seq2seq model:} which first generates a caption for the input image, and then, this caption is fed into a seq2seq model to generate a question, and (iii) \textbf{GRNN model:} where the CNN feature of the input image is passed to a gated recurrent unit (GRU) to generate questions.
We empirically observe that these methods often fall short in performance primarily due to their inability (a) to semantically interpret the text and visual content jointly, and (b) to establish consistency between generated question and the OCR token which is supposed to be the answer. To overcome these shortcomings, we propose a novel \prob{} model as described below. 
\newline
\newline
\noindent\textbf{Our Approach:} 
To encode visual content, scene text, its position and bringing consistency between generated question and OCR token; we propose an \underline{O}CR-consistent visual question generation model that \underline{L}ooks into the visual content, \underline{R}eads the scene text, and \underline{A}sks a relevant and meaningful natural language question. We refer to this architecture for \prob{} as \model{}. \model{} begins by representing visual features using pretrained CNN, extracted OCR token using FastText~\cite{fastText:2016} followed by an LSTM, and positions of extracted OCR using positional representations. 
Further, these representations are fused using a multimodal fusion scheme. The joint representation is further passed to a one-layered LSTM-based module and a maximum likelihood estimation-based loss is computed between generated and reference question. Moreover, to ensure that the generated question and corresponding OCR tokens are consistent with each other, we add an OCR token reconstruction loss that is computed by taking $l_2$-loss between the original OCR token feature and the representation obtained after passing joint feature to a multi-layer perception. The proposed model \model{} is trained in a multi-task learning paradigm where a weighted combination of the OCR-token reconstruction and question generation loss is minimized to generate a meaningful question. We evaluate the performance of \model{} on \prob{}, and compare against the baselines. \model{} significantly outperforms baseline methods on two public benchmarks, namely, ST-VQA~\cite{stvqa} and TextVQA~\cite{textVQA}.
\newline
\newline
\noindent\textbf{Contributions of this paper:} The major contributions of this paper are two folds:
\begin{enumerate}
    \item We draw the attention of the document image analysis community to the problem of text-based visual question generation by introducing a novel task referred to as \prob{}. \prob{} is an important and unexplored problem in the literature with potential downstream applications in building visual dialogue systems and augmenting training sets of text-based visual question answering models. We firmly believe that our work will boost ongoing research efforts~\cite{stvqa,docVQA:2021,textVQA,TextCaps:2020} in the broader area of conversational AI and text understanding. 

    \item We propose \model{} -- an OCR-consistent visual question generation model that looks into the visual content, reads the text and asks a meaningful and relevant question for addressing \prob{}. \model{} automatically generates questions similar to the datasets that are manually curated. Our model viz. \model{} significantly outperforms the baselines and achieves a BLEU score of 0.47 and 0.40 on ST-VQA~\cite{stvqa} and TextVQA~\cite{textVQA} datasets respectively.
\end{enumerate}

\section{Related Work}
\label{sec:relwork}
The performance of scene text recognition and understanding has significantly improved over the last decade~\cite{SceneText4:2015,SceneText5:2021,SceneText2:2012,SceneText6,SceneText3:2019,SceneText1:2011}. It has also started influencing other computer vision areas such as scene understanding~\cite{wordMatters}, cross-modal image retrieval~\cite{Mafla_2021_WACV}, image captioning~\cite{TextCaps:2020}, and visual question answering (VQA)~\cite{stvqa,docVQA:2021,textVQA}. Among these, text-based VQA works~\cite{stvqa,docVQA:2021,textVQA} in the literature can be considered one of the major steps by document image analysis community towards conversational AI. In the follow-up sections, we shall first review the text-based VQA followed by VQG which is the primary focus of this work.

\subsection{Text-based Visual Question Answering}
Traditionally, VQA works in the literature focus only on the visual content~\cite{first-vqa:2017,CLEVR:2017,prob-VQA:2020}, and ironically fall short in answering the questions that require reading the text in the image. Keeping the importance of text in the images for answering visual questions, researchers have started focusing on text-based VQA~\cite{stvqa,tVQA:2020,ocr-vqa,textVQA}. 
Among these works, authors in~\cite{textVQA} use top-down and bottom-up attention on text and visual objects to select an answer from the OCR token or a vocabulary. The scene text VQA model~\cite{stvqa} aims to answer the questions by performing reasoning over the text present in the natural scene. In OCR-VQA~\cite{ocr-vqa}, OCR is used to read the text in the book cover images, and a baseline VQA model is proposed for answering questions enquiring about author name, book title, book genre, publisher name, etc. Typically, these methods are based on a convolutional neural network (CNN) to perform the visual content analysis, a state-of-the-art text recognition engine for detecting and reading text, and a long short-term memory (LSTM) network to encode the questions.  
More recently, T-VQA~\cite{tVQA:2020} presents a progressive attention module and a multimodal reasoning graph for reading and reasoning. Transformer and graph neural network-based models have also started gaining popularity for addressing text-based VQA~\cite{MMGNN:2020,M4C:2020}. Another direction of work in text-based visual question answering is text-KVQA~\cite{textKVQA:2019} where authors propose a VQA model that reads textual content in the image, connects it with a knowledge graph, and performs reasoning using a gated graph neural network to arrive at an accurate answer.

\subsection{Visual Question Generation}
VQG is a dual task of VQA and is essential for building visual dialogue systems~\cite{diverse-vqg:2018,diverse2VQG:2017,vqg-infomax,dualVQG:2018,lbaVQG:2018,vqg0,diverse4VQG:2018,diverse3VQG:2017}. Moreover, the ability to generate relevant
question to the image is a core to in-depth visual understanding. Question generation from images as well as from raw text have been a well-studied problem in the literature~\cite{seq2seq:2017,diverse2VQG:2017,vqg-infomax,transformerVQA:2020,NLP-VQG4:2016,diverse4VQG:2018}. Automatic question generation techniques in NLP have also enabled chat-bots~\cite{seq2seq:2017,transformerVQA:2020,NLP-VQG4:2016}. 

Among visual question generation works, authors in~\cite{vqg0} focused on generating natural and engaging questions from the image, and provided three distinct datasets, each covering object to event-centric images. They proposed three generative models for tackling the task of VQG which we believe are the baselines for any novel VQG tasks, and we adopt these models for \prob{} and compare it with the proposed model. In~\cite{diverse-vqg:2018}, authors generated diverse questions of different types~\cite{diverse-vqg:2018}, such as, when, what, where, which, and how questions.
In~\cite{vqg-infomax}, a goal-driven variational auto-encoder model is used to generate questions by maximizing the mutual information between visual content as well as the expected answer category. In~\cite{dualVQG:2018}, authors posed VQG as a dual-task to VQA and jointly addressed both the task. 

These works in the literature restrict their scope to asking questions only about visual content and ignore any scene text present in the image. We argue that conversational agents must have the ability to ask questions by semantically bridging text in the image with visual content. Despite its fundamental and applied importance, the problem of \prob{} - text-based visual question generation has not been looked into in the literature. We fill this gap in the literature through our work.

\section{Proposed Model: \model}
\label{sec:approach}
Given a natural image containing text and an OCR token extracted from the image, \prob{} aims to generate a natural language question such that the answer is the OCR token.
To solve this problem, the proposed method should be able to successfully recognize text and visual content in the image, semantically bridge the textual and visual content and generate meaningful questions. For text detection and recognition, we rely on one of the successful modern scene text detection and recognition methods~\cite{Craft:2019,SceneText3:2019}.\footnote{For one of the datasets namely TextVQA, OCR-tokens extracted from Rosetta~\cite{rosetta} are provided with the dataset.} Further, the pre-trained CNN is used for computing the visual features and an LSTM is used to generate a question. The overall architecture of the proposed model viz. \model{} is illustrated in Fig.~\ref{fig:approach}. \model{} has the following three modules:

\begin{figure}[!t]
    \centering
    \includegraphics[width=1\linewidth]{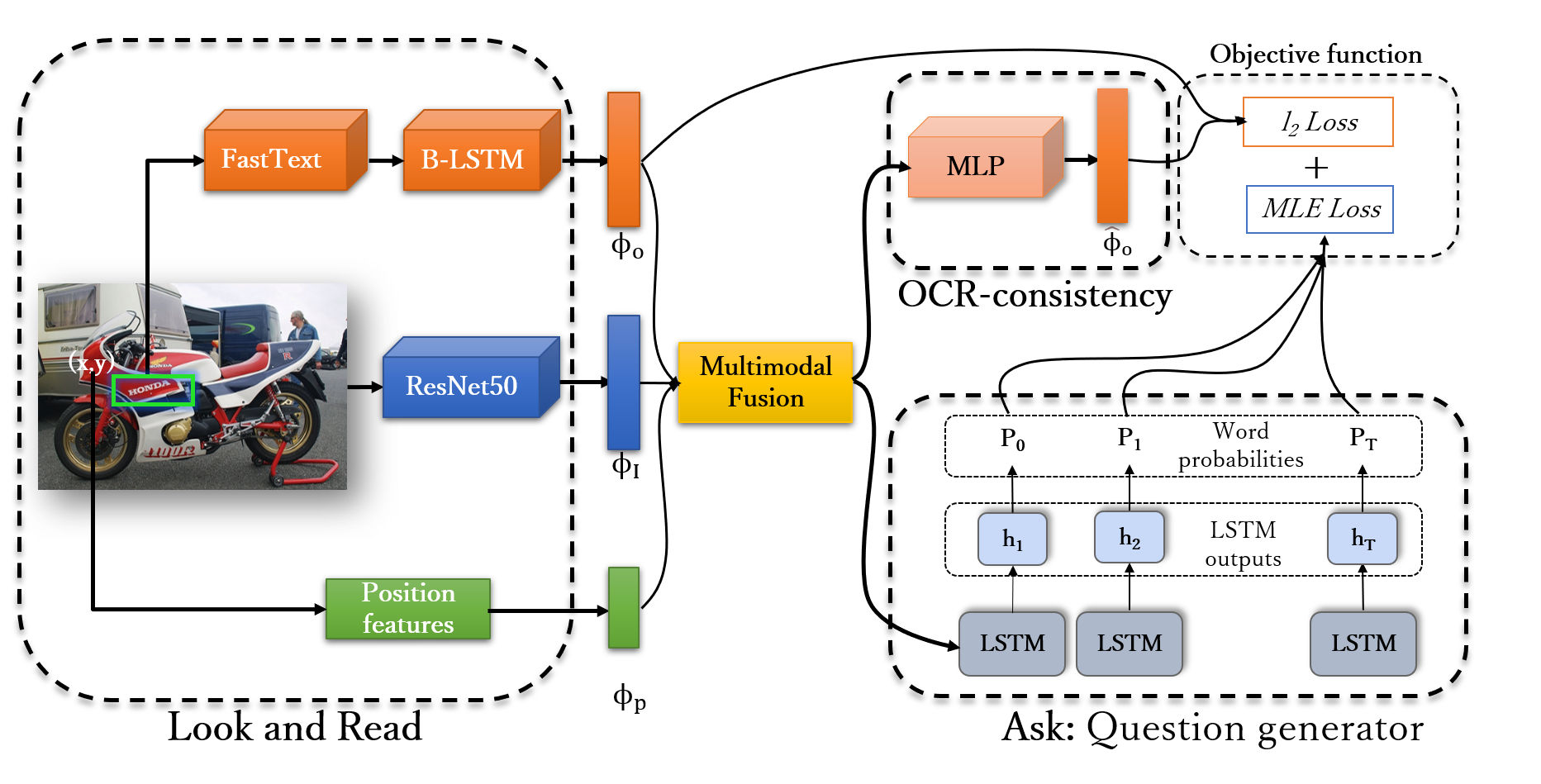}
    \caption{\label{fig:approach} \textbf{\model{}}. Our proposed visual question generator architecture viz. \model{} has three modules, namely, (i) Look and Read, (ii) OCR-consistency, and (iii) Ask module. Note: since OCR tokens can be more than a single word, we pass their FastText embeddings to a B-LSTM. Please refer to Section~\ref{sec:approach} for more details. \textbf{[Best viewed in color]}. }
\end{figure}
\noindent\textbf{(i) Look and Read Module:} In this module, we extract convolutional features from the given input image $I$. We use pre-trained CNN ResNet-50 which gives us 512-dimensional features $\phi_I$. Further, since our objective is to generate questions by relating visual content and text appearing in the image, we use~\cite{Craft:2019,SceneText3:2019} for detecting and recognizing text. We prefer to use~\cite{Craft:2019,SceneText3:2019} for text detection and recognition due to its empirical performance. However, any other scene text recognition module can be plugged in here. Once we detect and recognize the text, we obtain its FastText~\cite{fastText:2016} embedding. Since OCR tokens can be proper nouns or noisy, and therefore, can be out of vocabulary; we prefer FastText over Word2Vec~\cite{word2vec:2013} or Glove~\cite{glove:2014} word embeddings. The FastText embedding of the OCR token is fed to a bi-directional long short-term memory (B-LSTM) to obtain a 512-dimensional OCR-token feature or $\phi_o$.  

\begin{figure}[!t]
\begin{center}
 \includegraphics[width=1\linewidth]{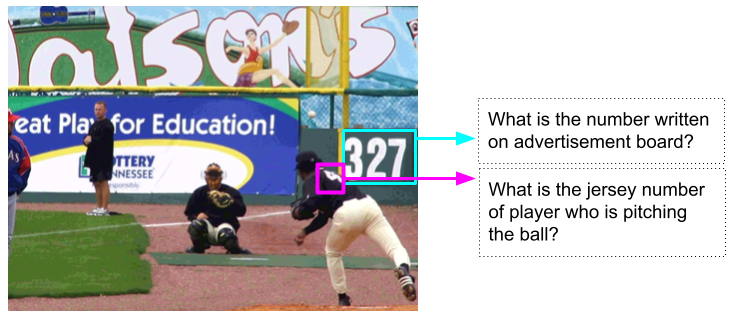}
\caption{\label{fig:pos} \textbf{Importance of positions of OCR tokens for question generation}. Numbers detected in different positions in the image i.e., number 4 and 327 in this example demands generating different questions.}
\end{center}
\end{figure}

Further, positions of OCR can play a vital role in visual question generation. For example as illustrated in Fig.~\ref{fig:pos}, in a sports scene, the number appearing in a sportsman's jersey and an advertisement board will require us to generate questions such as ``What is the jersey number of the player who is pitching the ball?" and ``What is the number written on advertisement board?" respectively. In order to use OCR token positions in our framework, we use 8 features i.e., topleft-x, topleft-y, width, height, rotation, yaw, roll, and pitch of the bounding box as $\phi_p$.\footnote{When we use CRAFT~\cite{Craft:2019} for text detection, we use only first four positional features i.e., topleft-x, topleft-y, width and height of the bounding box.}

Once these three features, i.e., $\phi_o$: OCR features, $\phi_I$: image features, and $\phi_p$: positions features are obtained, our next task is to learn joint representation by judiciously combining them. To this end, we use a multi-layer perceptron ${\cal F}$ to obtain joint representation $\Psi$ as follows:

\begin{equation}
    \Psi = {\cal F}(W, \phi).
\end{equation}
Here, $W$ is a learnable weight matrix and $\phi = [\phi_o;\phi_I;\phi_p]$ is concatenation of features. The joint feature obtained at this module $\Psi$ is a 512-dimensional vector that is fed to OCR token-reconstruction and question generator as described next.
\newline
\newline
\noindent\textbf{(ii) OCR-consistency Module:} 
After learning the joint representation of both image and OCR features, the model should be able to generate questions given specific tokens. To ensure this consistency, it is important that joint representation learned after fusion ($\Psi$) preserves the OCR representation ($\phi_o$). We reconstruct $\hat{\phi_o}$ by passing joint feature $\Phi$ to a multi-layer perceptron to obtain the reconstructed answer representation. We minimize the $l_2$-loss between the original answer and the reconstructed answer representations, i.e.,
\begin{equation}
    {\cal L}_a = ||\phi_o -\hat{\phi_o}||_2^2.
\end{equation}
\noindent\textbf{(iii) Ask Module:}
Our final module generates the relevant questions to the image. We use a decoder LSTM to generate the question q' from combined space as explained in the previous section. The minimization of maximum-likelihood error (MLE) between the generated question and the true question in the training set makes the model learn to generate appropriate questions. The {512}-dimensional joint feature vector serves as the initial state to an LSTM. We produce one word of the output question at a time using the LSTM. This happens until we encounter an end-of-sentence token. At each step during decoding, we feed in an embedding of the previously predicted word and predict the next word. The loss between generated and true questions guides the generator to generate better questions. The obtained joint feature $\Psi$ from the previous module acts as an input to the question generator. This question generator captures the time-varying characteristics and outputs questions related to the image and text present in the image. As mentioned above, we minimize the MLE objective ${\cal L}_{q}$ between generated question q' and ground truth question ${q}$. 
\newline
\noindent\textbf{Training:} \model{} is trained in a multitask learning paradigm for two tasks, i.e., OCR-token reconstruction and question generation. We use training set of text-based visual question answering datasets to train the model by combining both MLE loss ${\cal L}_{q}$ and OCR-token reconstruction loss ${\cal L}_{a}$, as follows:

\begin{equation}
    {\cal L} = {\cal L}_{q} + {\lambda}{{\cal L}_a},
    \label{eq:loss}
\end{equation}
where $\lambda$ is a hyperparameter which controls relative importance in optimizing these two losses, and ensures that the generated questions have better semantic structure with respect to the textual content along with the scene of the image.

\section{Experiments and Results}
\label{sec:expts}
In this section, we experimentally validate our proposed model for \prob{}. We first discuss  
the datasets and  performance measures in Section~\ref{sec:data} and Section~\ref{sec:pm} respectively. We then describe the baseline models for \prob{} in  Section~\ref{sec:base} followed by implementation details in Section~\ref{sec:imp}.
The quantitative and qualitative experimental results of the proposed model and the comparative analysis is provided in Section~\ref{sec:resNdis}. \subsection{Datasets}
\label{sec:data}
As this is the first work towards a visual question generation model that can read the text in the image, there is no dedicated dataset available for this task. We, therefore, make use of the two popular text-based VQA datasets, namely,  TextVQA~\cite{textVQA} and ST-VQA~\cite{stvqa}. Note that, unlike these datasets where originally the task is to answer a question about the image, our aim is to generate questions. In other words, given an image and an OCR token (often answers in these datasets), our proposed method learns to automatically generate questions similar to these manually curated datasets. A brief statistical description related to these datasets are as follows:

\begin{enumerate}
\item \textbf{ST-VQA}~\cite{stvqa}: consists of $23,038$ images with $31,791$ question-answer pairs obtained from different public datasets. A total of $16,063$ images with $22,162$ questions and $2,834$ images with $3,910$ questions are used for training and testing respectively, considering only the question-answer pairs having OCR tokens as their answers. 
\item \textbf{TextVQA}~\cite{textVQA}: comprises of $28,418$ images obtained from openImages~\cite{openimages} with $45,336$ questions. It also provides OCR tokens extracted from Rosetta~\cite{rosetta}. We use $21,953$ images with $25,786$ questions and  $3,166$ images with $3,702$ questions in all as training and testing set respectively, considering only the question-answer pairs having OCR tokens as their answers.
\end{enumerate}

\subsection{Performance metrics}
\label{sec:pm}
Following the text-generation literature~\cite{evalText:2020}, we use popular evaluation measures such as bilingual evaluation understudy (BLEU), recall-oriented understudy for gisting evaluation - longest common sub-sequence (ROUGE-L), and metrics for evaluation of translation with explicit ordering (METEOR).
The BLEU score compares $n$-grams of the generated question with the $n$-grams of the reference question and counts the number of matches. The ROUGE-L metric indicates similarity between two sequences based on the length of the longest common sub-sequence even though the sequences are not contiguous. The METEOR is based on the harmonic mean of uni-gram precision and recall and is considered a better performance measure in the text generation literature~\cite{meteor}. Higher values of all these performance measures imply better matching of generated questions with the reference questions. 
\subsection{Baseline models}
\label{sec:base}
Inspired by the VQG models in literature~\cite{vqg0}, we present three baseline models by adopting them for \prob{} task.

\noindent\textbf{Maximum Entropy Language Model (MELM):} Here, a set of OCR tokens extracted from the image along with a set of detected objects using FasterRCNN~\cite{faster_rcnn} are fed to a maximum entropy language model to generate a question. 

\noindent\textbf{Seq2seq model:} In this baseline, we first obtain caption of the image using a method proposed in~\cite{Discriminability:2018}. Then, the generated caption and the extracted OCR tokens from the images are passed to a seq2seq model~\cite{s2s:2014}. In seq2seq model, the encoder contains an embedding layer followed by an LSTM layer and for decoding, we use an LSTM layer followed by a dense layer. The seq2seq model is trained for question generation tasks using the training set of datasets described earlier.

\noindent\textbf{Gated Recurrent Neural Network (GRNN):} In this model, we obtain visual features using InceptionV3~\cite{RethinkingTI:2016}. This yields a feature vector of $1 \times 1 \times 4096$ dimensions that are then passed to a GRU to generate a question. We train GRU for question generation by keeping all the layers of InceptionV3 unchanged.



\subsection{Implementation details}
\label{sec:imp}
We train our proposed network using the Adam optimizer with a learning rate of $1e-4$, batch size of $32$, and a decay rate of $0.05$. The value of $\lambda$ in Equation~\ref{eq:loss} is set to $0.001$. The maximum length of the generated questions is set to $20$. We train the model for $10$ epochs. In multi-modal fusion, a two-layer attention network is used with feature sizes of $1032$ and $512$ respectively. The model is trained on a single NVIDIA Quadro P5000.

\begin{table*}[!t]
\setlength{\tabcolsep}{2pt}
\centering
\caption{\label{tab:table_comparison}\textbf{Comparison of \model{} with baselines.} We observe that the proposed model viz. \model{} clearly outperforms baselines on both the datasets for \prob{}.}

\begin{tabular}{c*{7}{c}c}
\hline
\multicolumn{1}{c|}{} & \multicolumn{3}{c|}{\textit{ST-VQA}~\cite{stvqa} dataset} & \multicolumn{3}{c}{\textit{TextVQA}~\cite{textVQA} dataset}\\
\hline

\multicolumn{1}{l|}{ Method} &   BLEU~~~&  METEOR~~~&  \multicolumn{1}{l|}{ ROUGE-L~~~}& BLEU~~~& METEOR~~~& ROUGE-L\\

\hline

\multicolumn{1}{l|}{MELM} & 0.34 & 0.12 & \multicolumn{1}{c|}{0.31} & 0.30 & 0.11 & 0.30\\

\multicolumn{1}{l|}{Seq2seq} & 0.29 & 0.12 & \multicolumn{1}{c|}{0.30} & 0.27 & 0.11 & 0.29\\

\multicolumn{1}{l|}{GRNN} & 0.36 & 0.12 & \multicolumn{1}{c|}{0.32} & 0.33 & 0.12 & 0.30\\

\hline



\multicolumn{1}{c|}{Ours (\model{})} 
& \textbf{0.47} & \textbf{0.17} & \multicolumn{1}{c|}{\textbf{0.46}} & \textbf{0.40} & \textbf{0.14} & \textbf{0.40}\\
\hline

\end{tabular}
\end{table*}

\subsection{Results and discussions}
\label{sec:resNdis}
\subsubsection{Quantitative Analysis.}
We evaluate the performance of our proposed model i.e., \model{}, and compare it against three baseline approaches, namely, MELM, seq2seq, and GRNN on ST-VQA and TextVQA datasets. The comparative result is shown in Table~\ref{tab:table_comparison} using performance measures discussed in Section~\ref{sec:pm}. Note that higher values for all these popularly used performance measures are considered superior.

Among the three baseline approaches, GRNN generates comparatively better questions. Our proposed model i.e., \model{} significantly outperforms all the baseline models. For example, on ST-VQA dataset, \model{} improves BLEU score by 0.11, METEOR score by 0.05, and ROUGE-L score by 0.14 as compared to the most competitive baseline i.e., GRNN. It should be noted that under these performance measures these gains are considered significant~\cite{evalText:2020}. We observe similar performance improvement on TextVQA dataset as well. 


\begin{figure}[!t]
    \centering
    \includegraphics[width=1\linewidth]{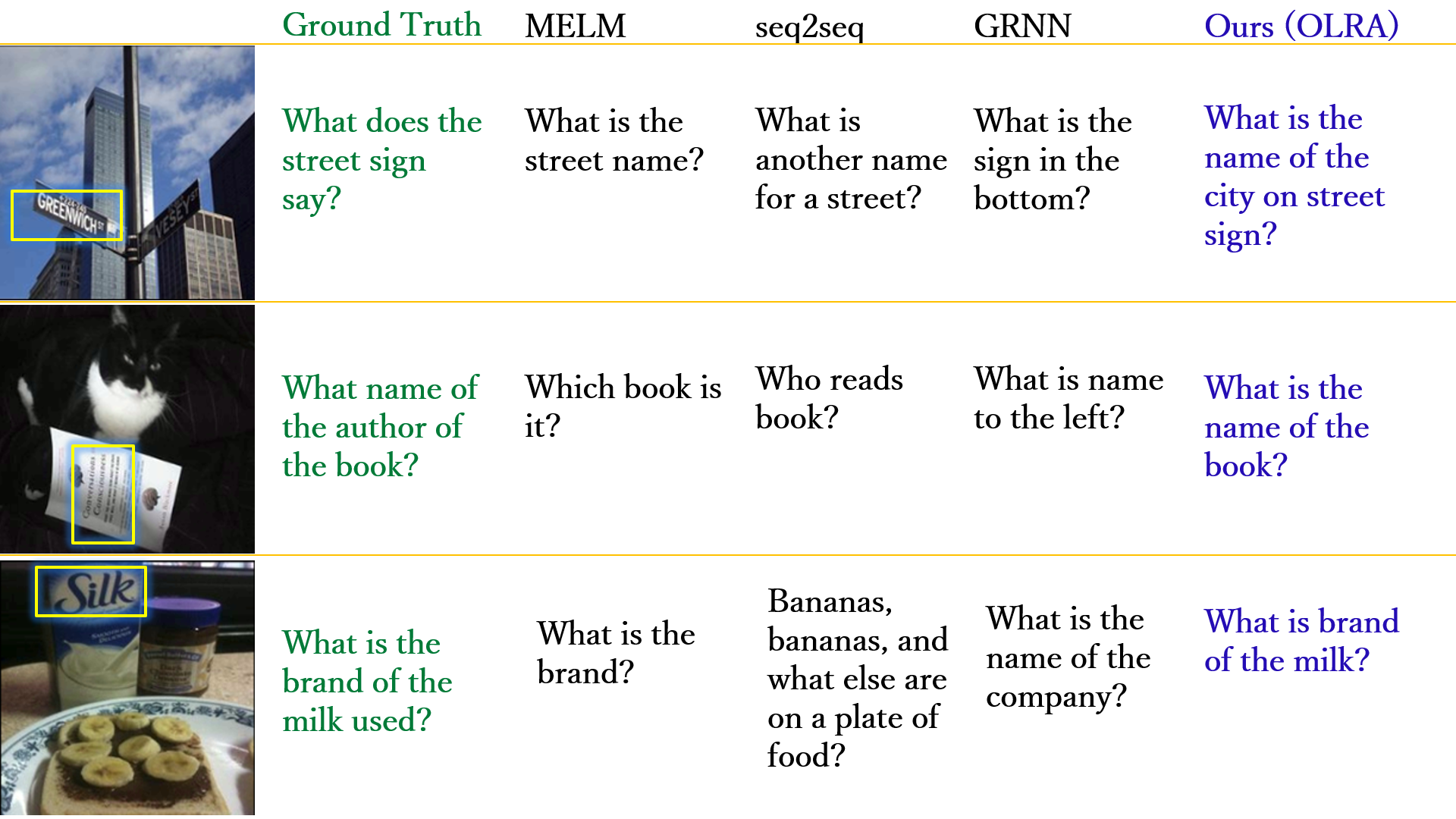}
    \caption{\label{fig:baseline_res} \textbf{Qualitative comparison.} Visual question generation by MELM, seq2seq, and GRNN baselines, and ours on a set of images from ST-VQA dataset.}
\end{figure}

\subsubsection{Qualitative Analysis.}
We perform a detailed qualitative analysis of the baselines as well as our proposed model. We first show a comparison of generated questions using all the three baselines versus \model{} in 
Fig.~\ref{fig:baseline_res}. We observe that the baselines are capable of generating linguistically meaningful questions. However, they do not fulfill the sole purpose of \textbf{Look}, \textbf{Read} and \textbf{Ask}-based question generation. For the first example in Fig.~\ref{fig:baseline_res} the expected question is ``What does the street sign say?", the baseline approaches and the proposed model generate nearly the same question. But, as the complexity of the scene increases, the baseline models fail to generate appropriate questions, and the proposed model due to its well-designed look, read, ask, and OCR-consistency modules, generates better questions. For example, in the last row of Fig.~\ref{fig:baseline_res}, MELM generates ``What is the brand?" but it fails to specify the kind of the product in the scene. Whereas the proposed model generates ``What is the brand of the milk?" which is very close to the target question.

\begin{figure}[h!]
    \centering
    \includegraphics[width=1\linewidth]{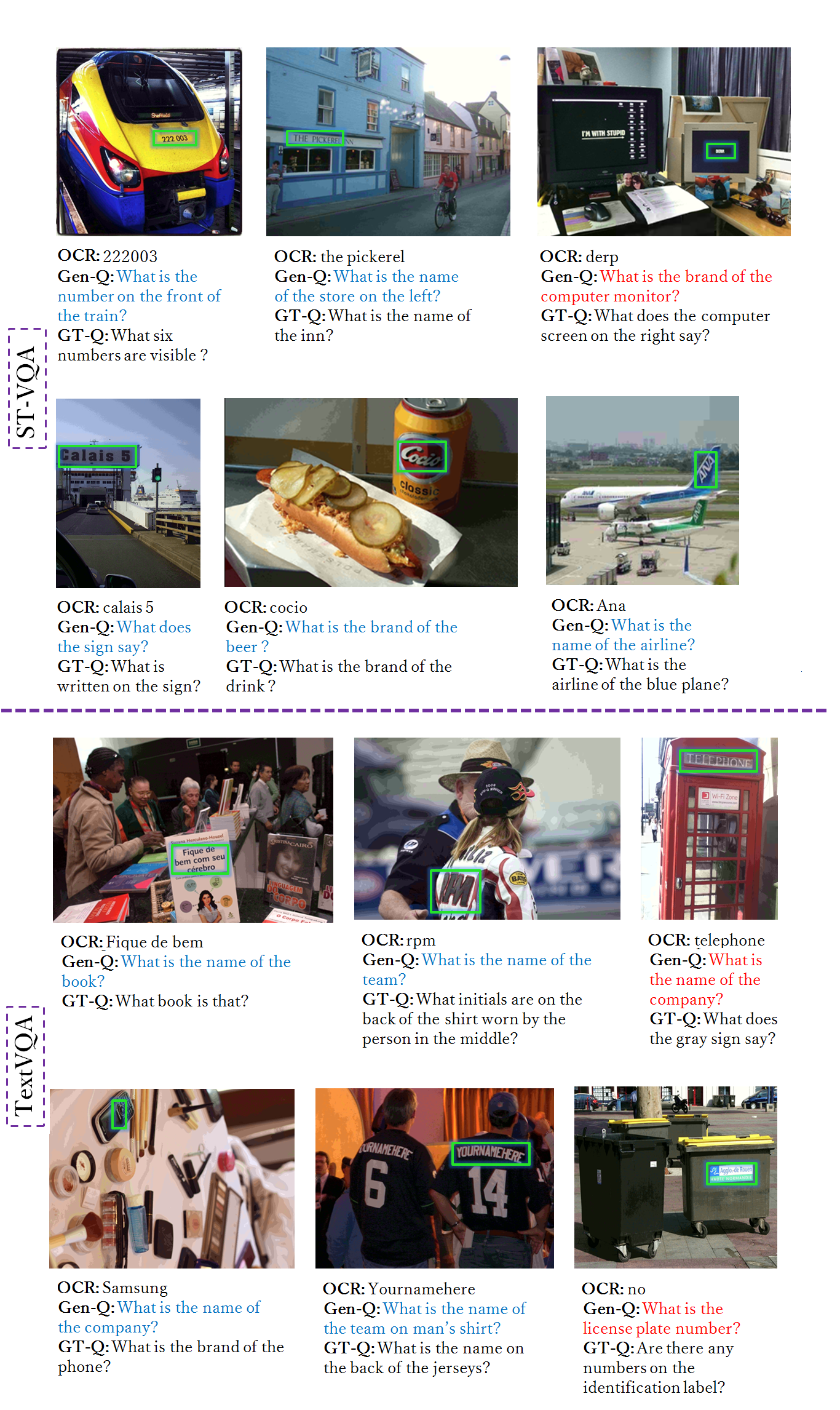}
    \caption{\label{fig:res_stvqa}Samples of generated questions on ST-VQA and TextVQA datasets with OCR tokens shown in \textcolor{green}{green} bounding box. Correctly and incorrectly generated questions are shown in \textcolor{blue}{blue} and \textcolor{red}{red} colors respectively. Gen-Q and GT-Q denote generated question from proposed method and ground-truth question respectively. \textbf{[Best viewed in color]}. }
\end{figure}


Further, more results of our model are shown in Fig.~\ref{fig:res_stvqa}. Here, we represent OCR token using green bounding boxes, correctly generated questions in the blue color text, and incorrectly generated questions in the red color text. Consider the example of a train image in Fig.~\ref{fig:res_stvqa}. Here, our method successfully generates the question ``What is the number on the front of the train?".  Similar such example question generations can be seen in Fig.~\ref{fig:res_stvqa}.

The failure of our model pronounced when either OCR-token is misinterpreted (for example, the word ``derp" on the computer screen is misinterpreted as computer monitor brand, and the word ``TELEPHONE" on a telephone booth is misinterpreted as company name)
or there is a need of generating questions whose answer is not an OCR token, for example: ``Are there any numbers on the identification label?". 

\begin{table*}[!t]
\setlength{\tabcolsep}{2pt}
\centering
\caption{\label{tab:ablation}\textbf{Ablation study.}  BLEU scores analysis (i) with inclusion of positional information and (ii) with n-word answers on both the datasets.}

\begin{tabular}{c*{3}{c}c}
\hline
\multicolumn{1}{c|}{~~~~~~~~OLRA} & \multicolumn{1}{|c|}{\textit{ST-VQA}~\cite{stvqa} dataset} & \multicolumn{1}{|c}{\textit{TextVQA}~\cite{textVQA} dataset}\\
\hline



\multicolumn{1}{c|}{~~~\textbf{w/o} position}
& {0.44}  & \multicolumn{1}{|c} {0.39} \\

\multicolumn{1}{c|}{~~~\textbf{w/} position}
& {0.45} &\multicolumn{1}{|c} {0.39} \\

\multicolumn{1}{c|}{~~~\textbf{w/} position and}
& {0.47}  &\multicolumn{1}{|c} {0.40} \\
\multicolumn{1}{c|}{~~~OCR-consistency}
& {}  & \multicolumn{1}{|c}  {} \\
\hline


\multicolumn{1}{c|}{~~~\textbf{w/} 1-word answers} 
& {0.48}  & \multicolumn{1}{|c}{0.41} \\

\multicolumn{1}{c|}{~~~\textbf{w/} 2-word answers} 
& {0.47} & \multicolumn{1}{|c} {0.39} \\

\multicolumn{1}{c|}{~~~\textbf{w/} 3-word answers} 
& {0.46} & \multicolumn{1}{|c}{0.40}  \\

\hline

\end{tabular}
\end{table*}

\noindent\textbf{Ablation Study:} 
We perform two ablation studies to demonstrate: (i) utility of the positional information and (ii) model's capability to generate those questions which have multi-word answers. 

Model without positional information considers image features $\phi_I$ + token features $\phi_o$, and with positional information considers image features $\phi_I$ + token features $\phi_o$ + positional information $\phi_p$ to generate questions. We observe that model with positional information and OCR-consistency i.e., our full model (\model{}) as shown in Table~\ref{tab:ablation} enhances the BLEU score and quality of generated question over other models on both ST-VQA and TextVQA datasets. 

Further, in Table~\ref{tab:ablation}, we also show \model{}'s performance on generating those questions that have 1-word, 2-word, and 3-word answers respectively. Based on statistical analysis with respect to ST-VQA test set, there are, $69\%$ 1-word, $19.6\%$ 2-word, $7\%$ 3-word, and $4.4\%$  above 3-word question-answer pairs. While in TextVQA test set, there are, $63.7\%$ 1-word, $21.3\%$ 2-word, $8\%$ 3-word, and $7\%$ above 3-word question-answer pairs. We observe that BLEU scores are nearly the same and the model performs equally well in all the three cases (see Table~\ref{tab:ablation}). This ablation study indicates that our model is capable of generating even those questions which have two or three-word length as the answer.

\section{Conclusions}
We introduced the novel task of `Text-based Visual Question Generation', where given an image containing text, the system is tasked with asking an appropriate question with respect to the OCR token. We proposed \model{} -- an OCR-consistent visual question generation model to ask meaningful and relevant visual questions. \model{} outperformed three baseline approaches on two public benchmark datasets. As the first work towards developing a visual question generation model that can read, we restrict our scope to generating simple questions whose answer is the OCR token itself. Generating complex questions that require deeper semantic and commonsense reasoning, and improving text-based VQA by augmenting its training data using automatically generated questions are few tasks that we leave as future works.  

We firmly believe that the captivating novel task and the benchmarks presented in this work will encourage researchers to develop better \prob{} models, and thereby gravitate ongoing research efforts of the document image analysis community towards conversational AI.

\bibliographystyle{splncs04}
\tiny{
\bibliography{refs} 
}
\end{document}